# Using MathML to Represent
# Units of Measurement
# for Improved Ontology Alignment *

Chau Do and Eric J. Pauwels

Centrum Wiskunde & Informatica, 1098 XG Amsterdam, The Netherlands
{do, eric.pauwels}@cwi.nl

**Abstract.** Ontologies provide a formal description of concepts and their relationships in a knowledge domain. The goal of ontology alignment is to identify semantically matching concepts and relationships across independently developed ontologies that purport to describe the same knowledge. In order to handle the widest possible class of ontologies, many alignment algorithms rely on terminological and structural methods, but the often fuzzy nature of concepts complicates the matching process. However, one area that should provide clear matching solutions due to its mathematical nature, is units of measurement. Several ontologies for units of measurement are available, but there has been no attempt to align them, notwithstanding the obvious importance for technical interoperability. We propose a general strategy to map these (and similar) ontologies by introducing MathML to accurately capture the semantic description of concepts specified therein. We provide mapping results for three ontologies, and show that our approach improves on lexical comparisons.

## 1   Introduction and Motivation

An increasing number of scientific and technological areas, including multi-agents, bioinformatics and the semantic web, are making use of ontologies to better represent their knowledge domain. An ontology describes a domain of interest by presenting a vocabulary as well as definitions of the terms used in the vocabulary [1]. With independent individuals and groups developing their own ontologies, we are faced with the problem of heterogeneous representation across ontologies. This is quite problematic when it becomes necessary to amalgamate or link data between various sources. Over the years, several solutions have been proposed for matching ontologies (i.e. identify corresponding or matching terms in different ontologies). Most take a generic approach in order to deal with the widest possible variety of ontologies from various domains. Consequently, these

---

matchers do not take advantage of domain specific attributes which could lead to better matches.

One area of application where a domain-agnostic approach might be sub-optimal concerns *units of measurement*. Units are particularly interesting since, unlike more common concepts which often carry multiple meanings, they have a clear definition due to an inherent mathematical structure. For example, *person* can be interpreted as either equivalent to *human* or as a subclass of *human*, both alignments are acceptable depending on the application. Contrast this to units of measurement, where there are well established rules, such that a unit defined in one ontology should only be matched to equivalent units in another ontology. Aligning units of measurement ontologies is of particular importance to areas requiring data sharing or conversion between units, just to name a few. For example, independent sensor networks may use different ontologies to represent their measurements and a mapping is required when their data is consolidated.

The aim of this paper is to propose a semi-automatic solution for the problem of aligning units of measurement ontologies. The solution we propose hinges on the use of MathML to extend the semantic description of the units that already exist in the ontology. To understand the underlying idea of our approach, consider two ontologies, $\Omega_1$ and $\Omega_2$, containing definitions for units of measurement. Assuming that these definitions contain both the dimensions and conversion values of the units, it cannot be assumed that the way this information is represented and encoded is similar. For example, assume in ontology $\Omega_1$ the unit *degree Celsius* is denoted as *degreeCelsius* respectively, whereas in ontology $\Omega_2$ it is known as *ThermoUnit_C*. More often than not, these two ontologies will have been developed by groups working independently. To circumvent this problem, we propose to insert for each unit in an ontology a MathML-encoded description using the information available in the ontology. In the *degree Celsius* example, a straightforward search based on a generic lexical comparison would find it difficult to spot this match. But if both ontologies contain the MathML-encoded relationship between Celsius and the base unit Kelvin, i.e. $T_c = T_k - 273.15$, then matching of these terms across the ontologies becomes trivial.

Our choice for using MathML in these encodings is motivated by the fact that this is already a widely accepted language for describing mathematical equations. Furthermore, due to its standardization, it is possible to write a generic program that can process different equations. Although in general, matching ontologies is difficult, the problem is made more manageable through the creation of a richer set of structures and relationships, which encode the precise mathematical relationships that exist between measurement units. This allows for more exact matchings as well as non-obvious ones (e.g. *NewtonPerMeter* and *joule_per_square_metre*).

This paper focuses on units defined in RDFS/OWL ontologies. For ontologies, even the ones within the confines of RDFS and OWL, there is no explicit requirement to represent the mathematical structure of units. The definitions vary from extremely minimal (for example only the names of the units) to more complex (some ontologies define dimensions, conversions, alternative symbols and so on). Even in the latter case, there is no clear and consistent manner for representing the mathematical structure. For example, to denote division between two units, one ontology defines the properties *numerator* and *denominator*, while another merely defines the property *hasOperand* and indicates the division by the inversion of the unit (e.g. *perKilogram*). To make matters worse, the labeling of concepts in ontologies are different (some examples encountered were: *cubic_metre* vs. *meterCubed*, *Vector_L1* vs. *SI_length_dimension_exponent* for the length dimension). Additionally, the structure and organization of concepts within ontologies can vary. Due to these variations, there is no logical link between units in different ontologies.

The semantic web is composed of layers, each building upon the previous one. The ontology vocabulary level defines the terms and relationships for concepts. The layer above this, the logic level, builds upon this foundation using reasoners to provide inferences. Reasoners lack the arithmetic skills to spot the correspondence between a statement such as "1 week *hasDuration* 7 days" in one ontology and the concatenation of statements "1 day *hasDuration* 86400 secs" and "1 week *hasDuration* 604800 secs" in another. Providing mathematical descriptions of these facts (encoded in MathML which is amenable to manipulations by software such as Mathematica) creates new opportunities for more effective identification.

The rest of the paper is structured as follows: first, in section 2 we describe some background information and related work. This is followed by a description of our proposed solution in section 3. Section 4, outlines the application of our approach to real life ontologies. Finally, section 5 outlines the results, followed by the conclusions and future work in section 6.

## 2 Background and Related Work

While this paper focuses on units represented within RDFS/OWL ontologies, units have been considered and represented in other related areas. OpenMath for example, deals with units using content dictionaries (CD). In [2] and [3], the representation of units in CDs is proposed and discussed. A question of whether or not RDF is a more suitable means of representation is also raised, but not definitively dealt with by the authors. [4] builds upon these CDs and suggests changes for better conformance to the SI standard. As will be explained later on, our method of inserting MathML into existing ontologies utilizes the information available within them. The information is extracted and MathML is automatically generated without any recognition of the unit that is being processed. Therefore we do not attempt to match the units to ones available in CDs.

An alignment between ontologies is described as a set of correspondences. Correspondences represent a relationship (equivalence, subclass, disjointness, etc.) between the entities of the two ontologies being aligned. Entities here can refer to classes, properties, individuals and so on. Consider two ontologies, $\Omega_1$ and $\Omega_2$, to be aligned, where $\Omega_1$ has the class *dog*, denoted here as $\Omega_1$:dog and

$\Omega_2$ has the classes *animal* and *canine*, denoted as $\Omega_2$:animal and $\Omega_2$:canine. The correspondences that make up the alignment would be:

1. $\Omega_1$:dog *is a subclass of* $\Omega_2$:animal
2. $\Omega_1$:dog *is equivalent to* $\Omega_2$:canine

The matching process can have additional inputs, such as a partial initial alignment, weights and thresholds (varies between matching algorithms) and sources for common knowledge (e.g. WordNet [5], UMLS [6]). Depending on the matching algorithm, the correspondence may have a level of confidence (normally between 0 and 1) associated with it [7].

Over the years, many ontology matching systems have been proposed, some of which are summarized in [1], [8], [9]. Although the approach taken by each system is different, most are based on terminological and structural methods. Terminological methods refers to the use of lexical comparisons of the labels, comments and/or other annotations of each entity. Structural comparisons look at for example similarities in the hierarchy of the ontology structure or the corresponding neighbors of matched entities. Semantic methods can also be applied for verification of matches or building on initial matches. These methods include looking at the range of values, cardinality, the transitivity and symmetry of the entities [7].

In an effort to find a common basis on which to compare ontology matching systems, the Ontology Alignment Evaluation Initiative (OAEI [10]) was formed. The initiative is composed of several tracks dealing with ontologies from different areas such as biomedical, conferences and anatomy. In particular the benchmark tests (see [11] for more information) have generated quantitative results, allowing for the comparison between different matching systems and tracking of advancements in these systems. It is clear that many of the systems are generic matchers, while some have more inclination towards specific areas (e.g. ASMOV towards the biomedical area). This is further indicated in the test case ontologies, which deal mainly with concepts from various domains.

Our proposed matching allows for an n to m cardinality (n entities in one ontology can align to m entities in the second ontology), which is an important improvement over a simple lexical comparison. Matching systems commonly only produce a one to one alignment [7]. Ones that provide an n to m alignment are AgreementMaker, COMA++ and ASMOV.

AgreementMaker comprises of a first layer, which produces similarity matrices based on concepts between the two provided ontologies. The features of the concepts (e.g. label, comments, annotations) are compared using syntactic and lexical comparisons. The results are fed into the second layer, which uses conceptual or structural methods to improve the results. Descendant's Similarity Inheritance (DSI) and Sibling's Similarity Contribution (SSC) are examples of the algorithms used for this stage. The last layer outputs a final matching or alignment by combining two or more matchers from the previous layers. For the first two layers, several matchers are available for comparison [12].

Similarly, COMA++ is based on an iterative process constructed of three main steps. The first is component identification, where relevant components for matching are determined. The second step is the matcher execution which applies multiple matchers in order to compute component similarities. The final step is similarity combination, where the correspondences between components is found from the calculated similarities [13].

ASMOV initially applies a pre-processing step to the two input ontologies. This step is terminological based and uses either an external thesaurus or string comparison method. The next step comprises of the structural methods, the first being a calculation of relation or hierarchical similarity. The second part comprises of internal or restriction similarity based on the established restrictions between classes and properties. Finally, an extensional similarity is found using data instances in the ontology [14].

Clearly these matchers are generically designed to deal with a wide variety of ontologies. Our approach focuses on the area of units of measurement and applies MathML to better represent the semantics of the units in order to increase correct equivalence alignments.

The incorporation of MathML into ontologies has been done before. For example, the Systems Biology Ontology (SBO) from the European Bioinformatics Institute [15] incorporates subject related equations using MathML. However, other than representing equations, the MathML is not being used further. More interesting usages of MathML can be seen in the Systems Biology of Microorganisms initiative, which has the aim of producing computerized mathematical models representing the dynamic molecular process of a micro-organism [16]. Within this initiative, SysMO Seek is an "assets catalogue" representing information such as models, experiments, and data. MathML is used to represent the mathematical models [17].

Another notable area where MathML and ontologies merge, is the OntoModel tool. Utilized for pharmaceutical product development, OntoModel allows for model creation, manipulation, querying and searching. It uses a combination of Content MathML and OWL. The former is used to represent the mathematical equations and the latter is used for the ontologies that represent the mathematical models and other related information [18].

While SysMo Seek and OntoModel use MathML to represent mathematical equations/models, the MathML is not used to align ontologies as we propose here.

## 3   Proposed Alignment Approach

The main contention of this paper is that MathML could play a pivotal role in this effort of efficient ontology alignment. MathML comes in two distinct flavors: *Presentation MathML* simply specifies what formulas should look like, while the aim of *Content MathML* is to encode the exact semantics of mathematical expressions. With the introduction of version 3.0, MathML has come closer in-line with OpenMath, particularly with respect to content dictionaries [19]. Obviously, we are interested in *Content MathML* and in the remainder of this paper we will use *MathML* as shorthand for *Content MathML version 3.0*.

Ontologies describing units of measurement routinely provide information on their fundamental physical dimensions (e.g. *length, mass, time,* etc. ) and their conversion value (e.g. *Celsius* = 5/9(*Fahrenheit* - 32)). As can be seen, the conversion value includes both a multiplier and an offset. A special case is dimensionless units, which are sometimes represented by an additional "dimension", and must be handled in a slightly different manner (more on this later). Usually, it is understood that the conversions convert back to the SI base and/or derived units. For example *watt* can be described as either *joule per second* ($W = J/s$) or *kilogram-meter-squared per second-cubed* ($W = kg \cdot m^2/s^3$). This wording in itself illustrates another problem with lexical representation. Does the term "squared" correspond to the meter only or to kilogram-meter? Although to a person, this is clear, to a machine different interpretations are possible, unless a convention has already been established. The introduction of MathML resolves this ambiguity. To determine if two units are equivalent, their dimensions and conversion values must match.

The basic idea underpinning our approach is very straightforward. Suppose we have two ontologies, say $\Omega_1$ which relates concepts $\alpha, \beta, \gamma, \ldots$ (we will denote this as $\Omega_1 = \{\alpha, \beta, \gamma, \ldots\}$) and a second ontology $\Omega_2 = \{\xi, \eta, \zeta, \ldots\}$. In addition, let us assume that we are given as prior knowledge that concept $\alpha$ in $\Omega_1$ is equivalent to concept $\xi$ (denoted here as $\Omega_1 : \alpha \leftrightarrow \Omega_2 : \xi$), as well as $\Omega_1 : \beta \leftrightarrow \Omega_2 : \eta$. If we now are able to determine (e.g. using MathML) that $\gamma = \alpha/\beta$ but also that $\zeta = \xi/\eta$ then we can confidently infer the previously unknown match $\Omega_1 : \gamma \leftrightarrow \Omega_2 : \zeta$.

Specifically, our matcher requires three inputs: $\Omega'_1$, $\Omega'_2$ and initial matchings (prior knowledge). $\Omega'_1$ and $\Omega'_2$ are ontologies, which have been modified by inserting MathML into them as an alternate representation of their units. In order to align the units in the two ontologies, the MathML representation is compared, with the initial matchings acting as a common reference point. Our matcher will provide correspondences with only equivalence relations. A more detailed explanation is given in the following sections.

### 3.1 Minimum Prior Knowledge

As pointed out earlier, all units can be described both in *SI base units* and in *derived units* which, in turn, can be re-expressed in base units. Therefore, it can be concluded that all units can be described using the seven SI base units: *meter, kilogram, kelvin, second, candela, ampere* and *mole*. In view of this, in order to match two unit ontologies, only the SI base units need to be matched as a starting point. This is used as the minimum prior knowledge that is required to process the MathML labels. An assumption that our general approach makes is that the unit conversion values are always relative to the SI base units and this is reflected in the MathML comparison. When two units of measurement ontologies are to be matched, the user must supply an initial matching between the base units found in each ontology.

### 3.2   Generation and Insertion of MathML

The difficulty in generating MathML and inserting it into an existing ontology depends on the structure of the ontology and what information is available in it. For instance, different ontologies will use different properties to indicate that one unit is the quotient of two other units. However, given a well structured ontology that is consistent in how it represents the units, a repeatable pattern will emerge as to where the necessary information for the MathML encoding can be found. A program can be written to automatically process these patterns and recast them in MathML code. We inserted MathML into three exisiting ontologies, one of which will be looked at in more detail later on. But in the end, the effort of inserting MathML will vary from ontology to ontology.

For the purposes of our approach, the MathML need only be inserted such that it is accessible by our matcher. Consequently, there is no need for integrating the MathML with the existing ontology such that any external semantic reasoner (e.g. Fact++, Pellet) can process it. With this in mind, we take a similar approach as OntoModel, which generates the MathML for an equation and incorporates it as a string into a *hasML* property [18]. While a specific data property could be developed, it did not make sense to create a new ontology just for one data property for the MathML code. Consequently, it was decided to incorporate the MathML into the ontologies as an `rdfs:comment` with an `rdf:parseType="Literal"` to indicate that markup language is being used (see [20] and [21]).

### 3.3   Processing of MathML

Once the ontologies have MathML inserted to represent their units, the matching process can begin to determine equivalent units. Before a comparison of the MathML code can be done, it must first be extracted from each ontology for every unit. By this, we mean that a search is done in each ontology for an `rdfs:comment` containing MathML code. The assumption by the matcher is that each unit that will be considered and aligned already has the corresponding MathML inserted. Expanding upon this initial matching of units (i.e. aligning entities without corresponding MathML) is a topic of future work (section 6). As noted previously, some units describe their conversions not in SI base units, but derived units. Both approaches are commonly used. Therefore, when comparing the MathML code, it must be checked to see if the units can be broken down further if they are not expressed in terms of base units.

As an example, consider we are given by the user the initial base units matchings of:

$$\Omega_1\text{:meter} \leftrightarrow \Omega_2\text{:metre}$$
$$\Omega_1\text{:kilogram} \leftrightarrow \Omega_2\text{:kilogram}$$
$$\Omega_1\text{:second} \leftrightarrow \Omega_2\text{:second\_time}$$
$$\Omega_1\text{:kelvin} \leftrightarrow \Omega_2\text{:kelvin}$$
$$\Omega_1\text{:candela} \leftrightarrow \Omega_2\text{:candela}$$
$$\Omega_1\text{:ampere} \leftrightarrow \Omega_2\text{:ampere}$$
$$\Omega_1\text{:mole} \leftrightarrow \Omega_2\text{:mole}$$

Where $\Omega_1$ is the first ontology and $\Omega_2$ is the second ontology. We encounter the units $\Omega_1$:joule and $\Omega_2$:newton\_metre during the matching process. They are given by the following equations (represented in MathML):

$$\Omega_1 : joule = (\Omega_1 : newton) \times (\Omega_1 : meter) \tag{1}$$

$$\Omega_2 : newton\_metre = \frac{(\Omega_2 : metre)^2 \times (\Omega_2 : kilogram)}{(\Omega_2 : second\_time)^2} \tag{2}$$

Here eq. (1) is expressed in the derived unit of newton, while eq. (2) is expressed completely in base units. Having only the base units as initial matchings, in order to compare these two units, the unit of $\Omega_1$:newton needs to be processed first. Therefore the following equation has to be first determined by the matcher:

$$\Omega_1 : newton = \frac{(\Omega_1 : meter) \times (\Omega_1 : kilogram)}{(\Omega_1 : second)^2} \tag{3}$$

Knowing eq. (3), when the matcher encounters eq. (1), it searches for $\Omega_1$:newton and upon finding it, reconstructs eq. (1) in its base units. Now the two units can be compared, with reference to the initial matchings. Once the dimensions and conversion values match, it can be concluded that the units are equivalent. This does not apply however to dimensionless units. For example, the units radian and steradian are both dimensionless and have a conversion multiplier of 1 and 0 offset. In this case, a lexical comparison (i.e. using different distance measurements) is used. When the comparison is completed, equivalence rules representing the alignment can be created and the results outputted to a file for later processing.

## 4   Application of Approach

As a proof of concept, the approach outlined in the previous section was applied to three ontologies. The implementation is divided into two phases.

f **Phase I** involves the following pre-processing steps for each individual ontology:

1. Find dimension and conversion data for the units
2. Generate MathML based on information of previous step and insert as `rdfs:comment`

3. Output modified file of ontology with the MathML code

**Phase II** compares two modified ontologies given initial matchings of the base units:

1. Read in the initial matchings file and two ontology files. Extract the MathML.
2. Compare the units (specifically their dimensions and conversion value) and determine which are equivalent
3. Output a file containing equivalence rules between the units of the two ontologies

The approach is broken down into two phases in order to make the implementation more modular . A general program can be written for Phase II, since the MathML is standardized. This program can be reused for comparison between any two units ontologies. The onus of inserting the MathML into the ontology can be placed on either its creator or a third party.

## 4.1   Phase I

**Inserting MathML into Existing Ontologies**   The three ontologies looked at in this work are: 1) Quantities, Units, Dimensions and Types (QUDT) [22], 2) Ontologies of units of measure 1.8 (OM) [23] and 3) Semantic Web for Earth and Environmental Terminology (SWEET) version 2.2 [24]. QUDT was originally developed by NASA for the NASA Exploration Initiatives Ontology Models project. It is currently being developed by TopQuadrant (see [25]) and NASA. The OM ontology was developed at Wageningen UR - Food & Biobased Research by the Intelligent Systems group. OM was designed to improve upon the deficiencies found in other units ontologies and was based on standards found in the field of units of measure. Additionally, the ontology is made more accessible by providing web services for things such as listing units by application area and unit conversion [26], [27]. SWEET is another ontology developed by NASA, but this time from the Jet Propulsion Laboratory. The focus of this ontology is on the Earth sciences and it bases its terms on the keywords found in the NASA Global Change Master Dictionary [28]. The three ontologies are supported by prominent organizations, while OM purports to be an improved ontology, designed in light of the flaws of previous units ontologies.

All three ontologies are fairly different in their structures and labeling. As a result different programs were written to insert the MathML into each ontology. However the general approach is similar in that patterns in the structure of the ontologies were first identified. A program was then written, using the Apache Jena library (`http://jena.apache.org/`) for Java and SPARQL queries, to utilize these patterns in order to extract the necessary information for the resulting MathML equation. Due to space restrictions, we provide a description of only the insertion of MathML into the OM ontology.

The OM ontology is well structured with units broken down into groups based on whether they are single, a multiple, an exponent or comprising of a division and so on. The units' mathematical relationship to other units is further expressed by object and data properties. To make this discussion more concrete, figure 1 shows an example of how unit division (in this case *millimetre per day*) is structured in the OM ontology. Here the unit division breaks down into the numerator and denominator object properties. In this example, they point to *millimetre* and *day* respectively. The two terms and their position in the division operation are clearly indicated through these properties. By following the numerator arm, it is seen that millimetre is comprised of the singular unit metre, which also happens to be a base unit. Millimetre also has a prefix *milli* with a value of 1e-3. This value comprises part of the overall conversion necessary to convert *millimetre per day* to *metre per second*. The denominator path breaks down the unit day into a numerical value of 86400 (the number of seconds in a day) and a unit of measure or measurement scale, second-time. The conversion value is determined as 1e-3/86400. The dimensions can be extracted in several ways. Either the speed-dimension can be directly processed or the length-dimension and time-dimension can be processed with the knowledge that one is the numerator and the other the denominator. Not shown in the figure, is that these dimensions break down into the seven dimensions: time, length, mass, amount of substance, temperature, electric-current and luminosity.

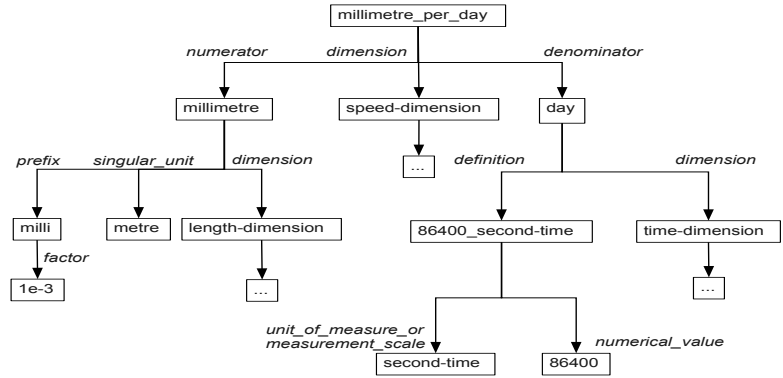

**Fig. 1.** Breakdown of a unit based on the division of two other units in the OM ontology. Length-dimension and time-dimension further break down into the basic seven dimensions: time, length, mass, amount of substance, temperature, electric-current and luminosity.

Other units consisting of a division, are represented similarly. The organization is different for unit multiplication, exponentiation and so on. What this example shows is that there are patterns in the OM structure which, once recognized, can be used to automatically determine the conversion value and dimensions of a unit. For example, numerical values found in the numerator section should be divided by the values found in the denominator section. In the case of multiplication where the unit breaks down into term 1 and term 2, values found after processing terms 1 and 2 should be multiplied. The dimensions are either processed directly if the unit has dimension data or constructed from the dimensions of the base units that comprise it.

After studying the OM ontology, we found that only a handful of these patterns exist. Recognizing this made it possible to write a program that searched the ontology, extracted the dimension data and calculated the conversion data.

The approach for the QUDT and SWEET ontologies was the same. In all three, we were able to identify patterns that covered the majority of units. Some units which did not fall within these patterns had to be handled manually. Reasons for this non-conformity vary from unusual units to errors in the ontologies.

**Generating MathML** Each unit is represented by an equation which incorporates its dimension and conversion data. In other words the unit is described in terms of its SI equivalent units and the conversion values necessary to convert to these SI units. The general structure of such a conversion equation is shown in eq. 4 below:

$$unit = a \times \frac{[n_1^{x1}][n_2^{x2}][n_3^{x3}]...}{[d_1^{y1}][d_2^{y2}][d_3^{y3}]...} + b \qquad (4)$$

Here $a$ represents the conversion multiplier and $b$ the conversion offset of the unit. The variables $n_i$ and $d_j$ represent the different units this unit is comprised of. So as noted before, the latter can be base SI units or derived units. Basically, for our approach to work they can be any other unit as long as it is possible to trace them back to a base SI unit. At least one of $n_i$ or $d_j$ should be present, but there is no limitation on the combination of these variables, this depends on the unit. The structure of the general conversion eq. 4 is fairly straightforward, simplifying the generation of the MathML encoding. An example of MathML code generated and inserted as a label is given in figure 2 for the unit newton.

As can be seen in the figure, the references to the other units in the ontology are given by the `id` attribute. The variables *n1*, *n2*, *d3* are equivalent to the $n_i$ and $d_j$ in eq. 4. In this manner, the variables show the relationship of one unit to other ones in the ontology, which can eventually be traced back to the SI base units. After the MathML is inserted into the `rdfs:comment` of each unit, the modified model of the ontology is outputted to a file.

### 4.2 Phase II

The implementation of this phase can be a standalone program. It will process ontology files containing MathML in their `rdfs:comment`. In addition, an initial alignment containing equivalences between the seven SI base units is provided to the program. Below is a detailed description of the steps.

**Extract MathML** A search through the ontologies for all individuals containing MathML code is initially done. This is done by conducting a SPARQL query for all `rdfs:comment` and a filter is applied for only comments containing

```
<math xmlns="http://www.w3.org/1998/Math/MathML">
    <bind><csymbol cd="fns1">lambda</csymbol>

        <bvar><ci id="myOntology:Meter">n1</ci></bvar>
        <bvar><ci id="myOntology:Kilogram">n2</ci></bvar>
        <bvar><ci id="myOntology:Second-Time">d3</ci></bvar>

        <apply><csymbol cd="arith1">divide</csymbol>

            <apply><csymbol cd="arith1">times</csymbol>
                <apply><csymbol cd="arith1">power</csymbol>
                    <ci xref="myOntology:Meter">n1</ci><cn>1</cn>
                </apply>
                <apply><csymbol cd="arith1">power</csymbol>
                    <ci xref="myOntology:Kilogram">n2</ci><cn>1</cn>
                </apply>
            </apply>

            <apply><csymbol cd="arith1">power</csymbol>
                <ci xref="myOntology:Second-Time">d3</ci><cn>2</cn>
            </apply>

        </apply>
    </bind>
</math>
```

**Fig. 2.** Sample MathML for unit *newton* (*N*). It encodes the fact that $N = (m \cdot kg)/s^2$ Notice how the SI base units are identified using the `id` attribute. The `xref` to the same unit in the $< ci >$ tag makes the relationship more explicit.

MathML. The results of this query are assumed to be all the processable units. In other words, anything without MathML is ignored (see Future Work, section 6). The MathML is then parsed to extract the dimension and conversion data.

**Compare Units** Once all the units in each ontology have the necessary information extracted, a comparison can be made using the initial matching data. Since no further information is known about the ontologies, a very general approach was taken. In the first pass, each unit in one ontology is compared to all the ones in the other. To compare the dimension data, the initial matching units and units that have already been found to be the same, are referred to. The reason for this is, as mentioned before in section 3.3, some units may be described in terms of other ones. Hence, a second pass is necessary to catch all the units which were not matched due to this reason. The steps of the comparison are summarized in figure 3.

- **Step 1:** First the simplest comparison is made by checking if the offsets of the conversion value are the same. If they are not, the units are not equivalent and a false is returned by the function.
- **Step 2:** Second, the multiplier of the conversion value is compared to see if they are the same.
- **Step 3:** Once the conversion value is confirmed to be the same, the dimensions are looked at next. If the units are expressed in units other than base SI units, these must first be broken down or matched. For example, *tesla* can be given as $T = N/(A \times m)$. If *tesla* is described in terms of *newton (N)* in both ontologies and *newton* has already been matched, then no breakdown is required. Otherwise a search is done for *newton* (already checked units are stored in memory) and if found, $T$ will be modified to $T = (kg \times m)/(A \times m \times s^2)$.
- **Step 4:** The next step, reduce dimensions, checks if there are the same units in the denominator and numerator and reduces them, resulting in $T = kg/(A \times s^2)$.
- **Step 5:** Once the units have been reduced as necessary, they can be compared with reference to the initial mappings and already matched units, to see if they are the same. If they are, the units match. Output the matched units.

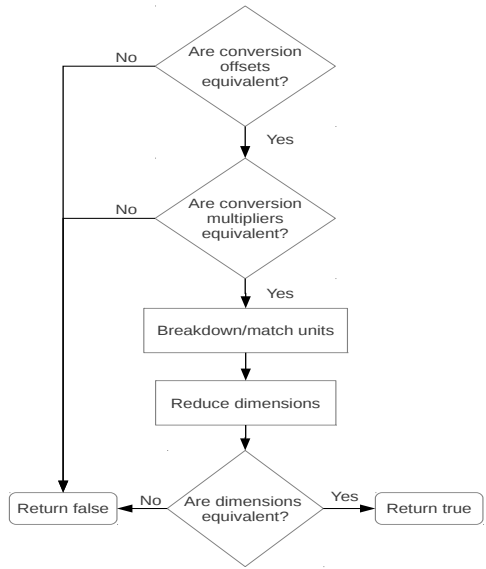

**Fig. 3.** The steps taken when comparing two units. First the conversion value is compared. If these are equal, the units (representing the dimensions) are matched or broken down into their base SI units if possible. Next, the reduce dimensions step checks if there are the same units in the denominator and numerator and reduces them. Finally, they are compared with reference to already matched units to determine if the dimensions are the same.

## 5 Results

To evaluate our matching approach, we manually aligned the ontologies for comparison (referred to as reference alignments). Following suit with the OAEI comparisons, we calculate the precision, recall and F-measure. The measurements of precision and recall are well known in information retrieval, but have been modified to take into consideration the semantics of alignments for the purposes of evaluating ontology alignments [29]. For this reason, we use the Alignment API version 4.4 to compare the generated alignments from our method with the reference alignments. The results are given in table 1.

The F-measures, being a combination of the precision and recall values, are fairly good. As a point of reference, the highest F-measure produced by the matchers participating in the OAEI competition from 2007-2010 was around 0.86 [1]. As can be seen the recall values are very good, indicating that most of the alignments in the reference are covered by the generated ones. The precision values are lower, indicating there are a number of false positives (i.e. units that were incorrectly identified as equivalent by the MathML approach). Looking closer at the results, the false positives fall into the following categories:

1. Mathematically equivalent but conceptually different units: There are two sub-types within this category. The first covers matches such as *hertz = becquerel*. While they are mathematically equivalent (both being equal to *1/s*), conceptually they are different, with the former representing frequency and the latter representing radioactive decay. The second sub-type encompasses matches such as *(square meter · steradian) = square meter*. When reduced completely, *steradian* becomes dimensionless and the equation is once again mathematically equivalent. This problem could be dealt with by modifying the *Reduce dimensions* step in the comparison. Both problems could also be handled by adding additional checks (e.g. lexical comparison of the labels).

2. Incorrect information in the ontologies: The insertion of the MathML is dependent on the information in the ontologies and if this information is incorrect, the resulting MathML and therefore the comparison is affected. Several problems were found in each of the ontologies. For example in QUDT, there are incorrect conversion values for the units of *teaspoon, tablespoon* and *centistokes*. Also there are no conversion values for the units of *degree Celsius per minute* and *year tropical*, to name a few. In the OM ontology the dimensions were wrong for the current density dimension and the permittivity dimension. In the SWEET ontology, some of the units were incorrectly composed. For example, the unit joule is only composed of *perSecondSquared* and *kilogram*, missing the *meter squared*.

**Table 1.** Precision, recall and F-measure values

| Alignment | Precision | Recall | F-Measure |
|-----------|-----------|--------|-----------|
| OM-QUDT | 0.81 | 0.95 | 0.87 |
| SWEET-QUDT | 0.77 | 0.97 | 0.86 |
| SWEET-OM | 0.82 | 0.99 | 0.90 |

These issues can be improved upon in future work, which will increase the precision values. Supporting documents for the results can be found at [30].

## 6 Conclusion and Future Work

Ontology alignment is a difficult problem, but by harnessing domain specific attributes, this problem can be simplified. We have shown that in the area of units of ontologies, MathML can be used to better represent the semantics of the units in order to compare them between ontologies. The generated alignments provide good precision and recall values when compared to manually created reference alignments.

For future work, we intend to improve upon the results by using further checks to ensure that the matched units are conceptually correct as well as mathematically. Furthermore, it will be interesting to look at combining this approach with other methods of ontology alignment. For example, the MathML matching can be used as an initial match in combination with lexical comparisons for non-mathematical concepts. This initial mapping is then fed into an algorithm which considers structural similarities between the two ontologies to build upon the initial matching. Another advantage of inserting MathML is that the information for conversion between units is more explicit. Instead of having to find the dimension information (to see if the units are compatible) and the conversion information within the ontologies, the MathML can be referred to. We intend to explore this area in the future for different applications, such as automatic unit conversion of sensor data between different networks.

*Acknowledgements* The authors gratefully acknowledge financial support by CWI's Computational Energy Systems project (sponsored by the Dutch National Science Foundation NWO) and wish to thank Dr. Christoph Lange for his valuable comments.